\documentclass[letterpaper]{article} 
\usepackage{aaai2026}  
\usepackage{times}  
\usepackage{helvet}  
\usepackage{courier}  
\usepackage[hyphens]{url}  
\usepackage{graphicx} 
\usepackage{natbib}  
\usepackage{caption} 
\usepackage{algorithm}
\usepackage{algorithmic}

\usepackage{amsmath,amsfonts,amssymb}
\usepackage{booktabs}
\usepackage{subcaption}
\usepackage{multirow}
\usepackage{pifont}

\usepackage{newfloat}
\usepackage{listings}
\DeclareCaptionStyle{ruled}{labelfont=normalfont,labelsep=colon,strut=off} 
\floatstyle{ruled}
\newfloat{listing}{tb}{lst}{}
\floatname{listing}{Listing}
\title{SkipCat: Rank-Maximized Low-Rank Compression of Large Language Models via Shared Projection and Block Skipping}
\author {
    Yu-Chen Lu\textsuperscript{\rm 1,2},
    Sheng-Feng Yu\textsuperscript{\rm 1,2},
    Hui-Hsien Weng\textsuperscript{\rm 1}, 
    Pei-Shuo Wang\textsuperscript{\rm 1}, \\
    Yu-Fang Hu\textsuperscript{\rm 1,3}, 
    Liang Hung-Chun\textsuperscript{\rm 3}, 
    Hung-Yueh Chiang\textsuperscript{\rm 4},
    Kai-Chiang Wu\textsuperscript{\rm 1}
}
\affiliations {
    \textsuperscript{\rm 1}National Yang Ming Chiao Tung University, \\
    \textsuperscript{\rm 2}Macronix International Co., Ltd., \\
    \textsuperscript{\rm 3}Skymizer Taiwan Inc., \\
    \textsuperscript{\rm 4}The University of Texas at Austin\\
    yuchen.cs11@nycu.edu.tw
}

\usepackage{bibentry}

\usepackage{xcolor}

\newcommand{\yuchen}[1]{{\color{black}#1}\normalfont}

\begin{document}

\maketitle

\begin{abstract}
Large language models (LLM) have achieved remarkable performance across a wide range of tasks. However, their substantial parameter sizes pose significant challenges for deployment on edge devices with limited computational and memory resources. Low-rank compression is a promising approach to address this issue, as it reduces both computational and memory costs, making LLM more suitable for resource-constrained environments. Nonetheless, naïve low-rank compression methods require a significant reduction in the retained rank to achieve meaningful memory and computation savings. For a low-rank model, the ranks need to be reduced by more than half to yield efficiency gains.
Such aggressive truncation, however, typically results in substantial performance degradation.
To address this trade-off, we propose \textit{SkipCat}, a novel low-rank compression framework that enables the use of higher ranks while achieving the same compression rates. First, we introduce an intra-layer shared low-rank projection method, where multiple matrices that share the same input use a common projection. This reduces redundancy and improves compression efficiency. Second, we propose a block skipping technique that omits computations and memory transfers for selected sub-blocks within the low-rank decomposition. These two techniques jointly enable our compressed model to retain more effective ranks under the same compression budget.
Experimental results show that, \textit{without any additional fine-tuning}, our method outperforms previous low-rank compression approaches by 7\% accuracy improvement on zero-shot tasks under the same compression rate. These results highlight the effectiveness of our rank-maximized compression strategy in preserving model performance under tight resource constraints.
\end{abstract}


\section{Introduction}

\begin{figure}[!t]
  \centering
  \includegraphics[width=\linewidth]{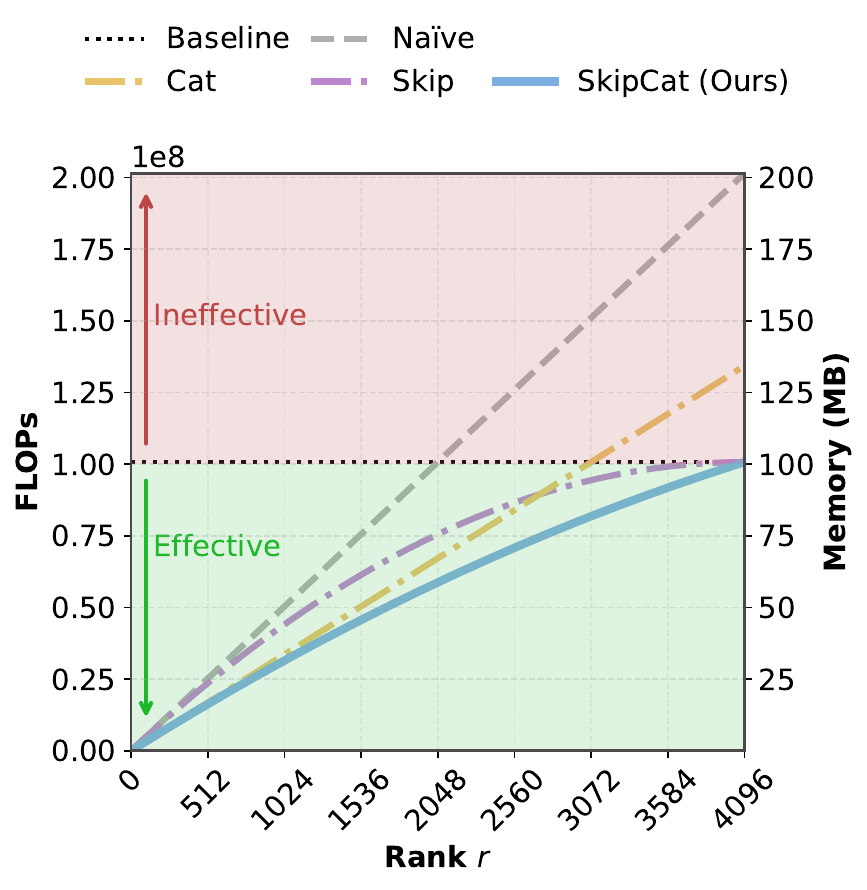}
  \caption{Computational cost (FLOPs) and memory footprint (MB) of the LLaMA2-7B attention module versus retained rank \(r\). The black dotted line represents the original computational cost and memory footprint of the module. The gray dashed line corresponds to a naïve SVD compression, which only achieves effective compression when the retained rank is reduced to less than half of the full rank. In contrast, \textit{SkipCat} maximizes the ranks and reduces the computational and memory cost, resulting in better trade-offs between efficiency and performance.}

  \label{fig:flops}
\end{figure}

Large language models (LLM) \cite{qwen3, touvron2023llama} have been employed in a wide range of real-world scenarios, such as professional problem-solving and intelligent home devices. However, deploying LLMs on edge devices remains challenging. For instance, the large number of parameters may exceed the limited memory capacity of such devices. In addition, their high computational cost can lead to excessive latency and energy consumption, which limits their practical use in resource-constrained environments. Consequently, substantial research \cite{yuan2024llm} efforts have been devoted to enhancing the efficiency of LLM inference.

Model compression \cite{zhu2024survey} has become a widely used strategy to reduce both model size and computational cost. Quantization techniques achieve this by lowering the precision of both model parameters and activations \cite{frantar-gptq, liu2024spinquant}, thereby reducing memory transfer overhead and potentially improving computational efficiency. Model pruning \cite{akhauri2024shadowllmpredictorbasedcontextualsparsity, sun2023wanda} improves inference efficiency by removing redundant or less important parameters.

\begin{figure*}[h]
  \centering
  \includegraphics[width=\textwidth]{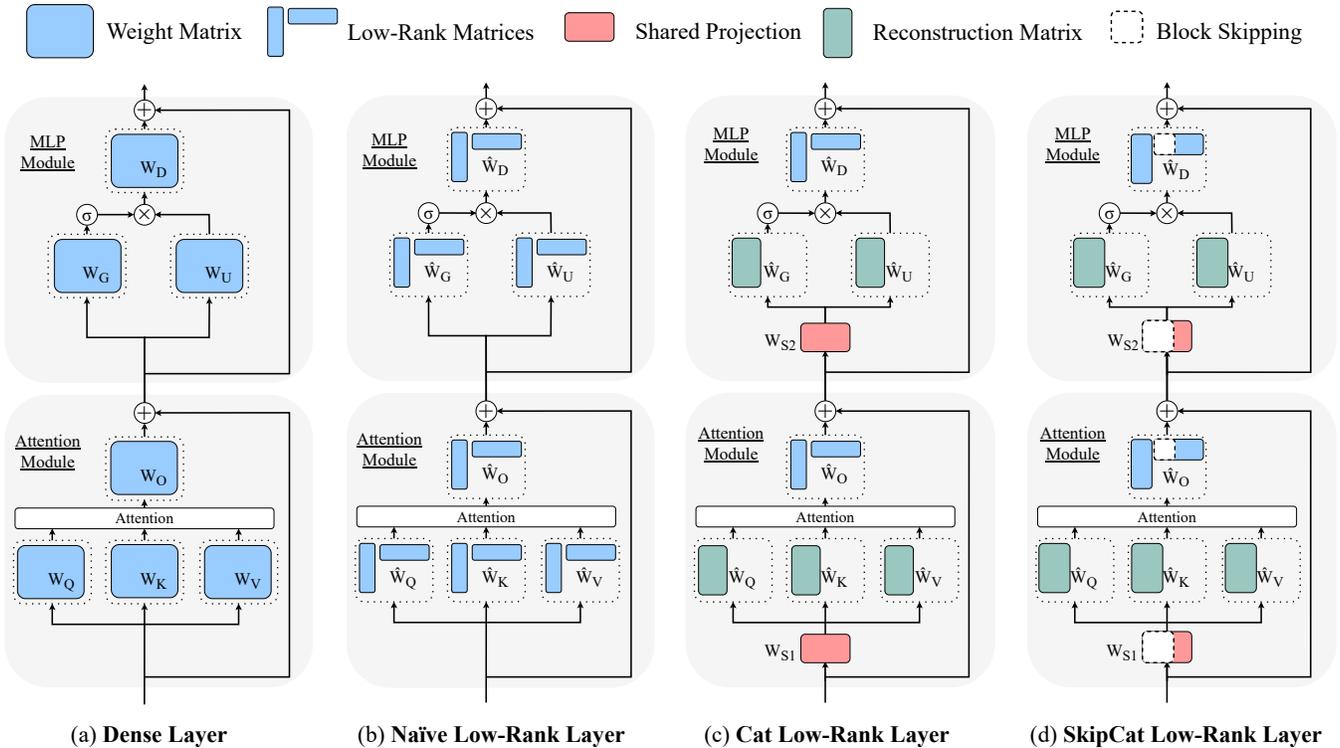}
  \caption{An illustration comparing our proposed method \textit{SkipCat} with naïve low-rank compression. (a) The original model layer. (b) The model structure after applying naïve SVD low-rank compression. (c) Our \textit{Cat} technique, which shares low-rank projection matrices across modules, enabling more ranks to be preserved without changing the compression rate. (d) Our \textit{Skip} technique, applied on top of the \textit{Cat} architecture, skips submatrix computations for all low-rank projections and further increases the number of preserved ranks.}
  \label{fig:model}
\end{figure*}

In addition to the aforementioned techniques, low-rank compression \cite{yuan2023asvd, wang2025svdllm} has also emerged as a promising research direction. By factorizing a weight matrix into two matrices with a shared low rank, we can reduce the number of parameters. 
Lowering the rank increases the compression rate by reducing the number of parameters and operations, but this may come at the cost of reduced accuracy.
Conversely, a higher rank allows the model to better preserve its original performance, at the cost of a larger model size. A well-known limitation of existing low-rank compression methods often requires a substantial decrease in ranks to obtain efficiency gains, which can significantly impair model performance. As illustrated in Figure~\ref{fig:flops}, taking the attention module of LLaMA2-7B as an example, the existing methods only begin to reduce floating-point operations (FLOPs) and memory below that of the original module when the retained rank falls below half of the original value (i.e., full rank). An excessive reduction can result in a significant drop in accuracy. Motivated by this observation, we aim to develop a method that increases the number of retained ranks while achieve the same compression rates, thereby mitigating the loss in model performance. To this end, we propose \textit{SkipCat}, a low-rank compression method that maximizes the preserved ranks than existing approaches under the same compression rate, thereby allowing the model to retain better performance after compression.


Unlike naïve SVD methods that decompose each weight matrix separately, we propose \textit{Intra-Layer Shared Low-Rank Projection using Matrix Concatenation} (\textit{Cat}), as illustrated in Figure~\ref{fig:model}(c). This technique constitutes one of the core components of our framework. In \textit{Cat}, multiple matrices with the same input share a single projection matrix, reducing the number of projection components and enabling the model to retain a greater number of ranks under the same compression budget. Another key technique in our approach is \textit{Block Skipping} (\textit{Skip}). As illustrated in Figure~\ref{fig:model}(d), \textit{Skip} enables the omission of sub-block computations across all low-rank projections, allowing additional ranks to be preserved without increasing the compression budget. Our method, \textit{SkipCat}, substantially improves rank retention and achieves highly effective compression rates, as shown in the blue curve in Figure~\ref{fig:flops}.

We evaluate our method on both the LLaMA \cite{touvron2023llama} and Qwen \cite{qwen3} models. At an equivalent compression rate, \textit{SkipCat} outperforms existing low-rank compression methods in the \textit{training-free} setting, achieving lower perplexity and up to a 7\% improvement in zero-shot task accuracy. 
These results demonstrate that both of our techniques (i.e., \textit{Skip} and \textit{Cat}) enable the maximization of effective ranks while maintaining efficiency gains and delivering superior model performance at higher compression levels.
We further evaluate \textit{SkipCat} on larger-scale models and in conjunction with LoRA fine-tuning techniques. The consistently strong results underscore the generalizability of our approach. Our approach offers a practical and effective solution for compressing LLMs, making them more feasible for edge devices. Our findings open new possibilities for future work in efficient model compression and deployment of large models in constrained environments.
Our main contributions are summarized as follows:
\begin{itemize}
    \item We propose a novel low-rank compression framework aimed at maximizing the number of effective ranks. The framework introduces two key techniques:
    (1) \textit{Cat}, which enables multiple matrices to share a common low-rank projection; and
    (2) \textit{Skip}, which allows the model to skip the computation of certain submatrices, while also addressing numerical instability issues that arise in half-precision inference.
    \item Experimental results show that our method consistently outperforms existing low-rank compression approaches in terms of compression performance.
\end{itemize}


\section{Related Works}

Numerous studies have explored techniques to reduce the computational and memory overhead of large language model inference. One major line of research focuses on lowering numerical precision, known as quantization. GPTQ \cite{frantar-gptq} is a state-of-the-art weight-only quantization method that converts model weights to low-precision formats, thereby reducing memory transfer costs during inference. Other works address weight-activation quantization, where the main challenge lies in handling activation outliers, which can severely degrade quantization accuracy. To address this, methods such as SmoothQuant \cite{xiao2023smoothquant} and SpinQuant \cite{liu2024spinquant} apply different scaling strategies to mitigate the impact of outliers, effectively preserving model performance after quantization. When supported by hardware, activation quantization can also lead to faster inference by enabling low-precision arithmetic throughout the computation.

In addition to reducing bit-widths, another major line of research focuses on minimizing the number of parameters to improve both memory and computation efficiency. Pruning techniques \cite{sun2023wanda, frantar-sparsegpt} aim to remove redundant and less important parameters from the model. While unstructured pruning can achieve high sparsity, it typically requires specialized hardware to realize performance gains. As a result, structured pruning \cite{ma2023llmpruner}, which removes entire units such as attention heads or channels, is more hardware-friendly.

Low-rank compression is another promising approach for reducing the number of model parameters. A common approach employs singular value decomposition (SVD) to approximate weight matrices in a lower-dimensional space. In the context of LLMs, several works have explored SVD-based techniques. LORD \citep{kaushal2023lord} applies SVD to compress code-focused LLMs, though directly applying naïve SVD to LLMs can lead to significant performance degradation. To mitigate such losses, FWSVD \citep{hsu2022language} leverages gradient information to assess each parameter’s importance, thereby minimizing the impact of compression on model accuracy. On the other hand, the presence of outliers in LLM activations not only hinders the effectiveness of quantization, but also poses a significant challenge for low-rank compression. ASVD \cite{yuan2023asvd} identifies this issue and proposes to mitigate the impact of outliers by applying a scaling matrix, derived from the activation distribution on calibration data. LatentLLM \cite{koike2025latentllm} \yuchen{enhances compression performance through the use of junction matrices. However, directly applying these matrices may lead to numerical instability.} SVD-LLM \cite{wang2025svdllm, wang2025svd} applies a whitening transformation to the parameter matrices as a preprocessing step, making the performance degradation approximately proportional to the discarded singular values. This leads to better performance compared to directly applying SVD without whitening. In contrast to prior methods, our proposed approach aims to increase the number of retained ranks without altering the overall compression rate, and simultaneously resolves the numerical instability that arises during low-precision computation.

Low-rank compression is also compatible with other techniques and can be integrated to achieve greater efficiency gains. Dobi-SVD \cite{qinsi2025dobisvd} integrates quantization with low-rank compression to achieve higher compression ratios without significant loss in model performance. Basis Sharing \cite{wang2024basis} shares the basis matrix among low-rank modules of the same projection type in different layers. While the above-mentioned method helps reduce the model's storage footprint in main memory, it brings no benefit in terms of computational cost or memory transfer during inference. In contrast, our method shares the projection matrix within \textit{the same layer} for matrices that operate on the same input. This design not only reduces memory transfer overhead but also alleviates computational cost.
\section{Proposed Method}

In this section, we first introduce the background of SVD-based compression in Section~\ref{subsec:svd}. We then present the two core components of our method in Sections~\ref{subsec:cat} and \ref{subsec:skip}. Finally, Section~\ref{subsec:skipcat} integrates these components into a unified framework and discusses the resulting benefits.

\subsection{SVD-based Low-Rank Compression}
\label{subsec:svd}

The typical layer structure of LLM, comprising attention and MLP modules, is illustrated in Figure~\ref{fig:model}(a).
\yuchen{For LLM, reducing the FLOPs of matrix computations and the size of weight matrices can help decrease computation time \cite{kaplan2020scaling} or memory transfer latency during both the prefilling and decoding stages.}
Given a weight matrix \( W \in \mathbb{R}^{d_{\mathrm{out}} \times d_{\mathrm{in}}} \), and an input vector \( x \in \mathbb{R}^{d_{\mathrm{in}} \times 1} \), the output is
\(
y = Wx \in \mathbb{R}^{d_{\mathrm{out}} \times 1}
\). \yuchen{We assume a batch size of 1 for simplicity.}
Each of the \( d_{\mathrm{out}} \) elements in \( y \) requires \( d_{\mathrm{in}} \) multiplications and \( d_{\mathrm{in}} \) additions, leading to a total of \( 2 d_{\mathrm{in}} d_{\mathrm{out}} \) FLOPs.
Additionally, storing the weight matrix requires \( d_{\mathrm{in}} d_{\mathrm{out}} \). 
This setting corresponds to the baseline case, as indicated by the black dotted line in Figure~\ref{fig:flops}.

Low-rank compression approximates large weight matrices with the product of smaller matrices, leading to reductions in both memory consumption and computation. The na\"ive approach is typically grounded in singular value decomposition (SVD), which factorizes a matrix into a set of orthogonal components. The weight matrix \( W \in \mathbb{R}^{d_{\mathrm{out}} \times d_{\mathrm{in}}} \) 
can be decomposed using SVD as follows:
\[
W = U \Sigma V^\top,
\]
where \( U \in \mathbb{R}^{d_{\mathrm{out}} \times d_{\mathrm{out}}} \) and \( V \in \mathbb{R}^{d_{\mathrm{in}} \times d_{\mathrm{in}}} \) are orthogonal matrices.
The matrix \( \Sigma \in \mathbb{R}^{d_{\mathrm{out}} \times d_{\mathrm{in}}} \) is a rectangular diagonal matrix, i.e.  the top-left \( R \times R \) block is a diagonal matrix, where \( R = \operatorname{rank}(W) \le \min (d_\mathrm{in}, d_\mathrm{out})\), and its nonzero diagonal entries \( \sigma_1, \sigma_2, \dots, \sigma_R \) are the  singular values of \( W \), arranged in non-increasing order:
\[
\sigma_1 \geq \sigma_2 \geq \dots \geq \sigma_R > 0.
\]
Each singular value indicates how much its associated component contributes to reconstructing the original matrix. Therefore, to achieve the best approximation at a given compression rate, one can retain only the top-\( r \) singular values with \( r \le R \) and their corresponding components from the decomposition.
This yields a truncated diagonal matrix \( \Sigma_r \in \mathbb{R}^{r \times r} \) containing the leading singular values. Simultaneously, we retain only the first \( r \) columns of \( U \) and  \( V \), resulting in the truncated matrices \( U_r \in \mathbb{R}^{d_{\mathrm{out}} \times r} \) and \( V_r \in \mathbb{R}^{d_{\mathrm{in}} \times r} \), respectively. The matrix \( W \) can then be approximated by a rank-\( r \) reconstruction:
\[
W \approx U_r \Sigma_r V_r^\top.
\]

To reduce computational cost during LLM inference, the weight-input multiplication can be approximated using the rank-\( r \) decomposition. By absorbing the singular values into the orthogonal matrices, the weight matrix can be expressed as the product of two smaller matrices, without increasing computational overhead, that is
%
\begin{equation}
W x \approx B A x, \label{eq:lrapprox}
\end{equation}
where \( B = U_r \Sigma_r^{1/2} \in \mathbb{R}^{d_{\text{out}} \times r} \) and \( A = \Sigma_r^{1/2} V_r^\top \in \mathbb{R}^{r \times d_{\text{in}}} \). \yuchen{We define \(A\) as the projection matrix and \(B\) as the reconstruction matrix in the low-rank decomposition, where \(A\) projects the input into a lower-dimensional space and \(B\) reconstructs the output.} Existing low-rank compression methods \cite{wang2025svdllm, wang2024basis} typically decompose the major weight matrices within each layer of the LLM, as illustrated in Figure 2(b).

\yuchen{After applying low-rank compression with rank \( r \), the total number of parameters in matrices \(A\) and \(B\) is \( r (d_{\mathrm{in}} + d_{\mathrm{out}}) \), with a computational cost of \( 2r(d_{\mathrm{in}} + d_{\mathrm{out}})\) FLOPs.}
Consequently, low-rank compression leads to a computational gain only when the reduced FLOPs and parameters are fewer than the original matrix multiplication. This condition yields the constraint:
\begin{equation}
r < d_{\mathrm{in}} d_{\mathrm{out}} \, / \, ({d_{\mathrm{in}} + d_{\mathrm{out}}}). \label{eq:lrconstraint}
\end{equation}

Equation~\eqref{eq:lrconstraint} highlights the limitation under which low-rank compression begins to yield computational benefits.
For example, if the weight matrix \( W\) is a full-rank square matrix, i.e. \( R=d_\mathrm{in}=d_\mathrm{out} \). The na\"ive low-rank compression  becomes efficient only when the rank \( r \) is less than half the full rank \( R \), that is \( r < R / 2 \), as illustrated by the gray dashed line in Figure~\ref{fig:flops}. This strong constraint motivates our work, which explores alternative approaches to relax this limitation and enable more efficient compression.

\subsection{Cat: Intra-Layer Shared Low-Rank Projection using Matrix Concatenation}
\label{subsec:cat}
As shown in the model architecture in Figure~\ref{fig:model}(b), we observe that in the attention module, \textit{q}, \textit{k}, and \textit{v} matrices share the same input, while in the MLP module, \textit{gate} and \textit{up} matrices also receive the same input. Therefore, if we allow matrices with identical inputs to share a single low-rank projection, as illustrated by the red components in Figure~\ref{fig:model}(c), we can increase the number of retained ranks without changing the overall compression ratio. To enable multiple modules to share the same low-rank projection, we apply \textit{concatenated} SVD, thereby obtaining a shared low-rank projection. 
We illustrate the following examples using the three matrices in the attention module, namely \(W_\mathrm{Q}\), \(W_\mathrm{K}\), and \(W_\mathrm{V}\), each in \(\mathbb{R}^{d_\mathrm{out} \times d_\mathrm{in}}\).
We first concatenate these matrices along the output dimension, as follows: 
\begin{align}
W_\mathrm{QKV} &= \begin{bmatrix} W_\mathrm{Q}^\top & W_\mathrm{K}^\top & W_\mathrm{V}^\top \end{bmatrix}^\top \in \mathbb{R}^{3d_\mathrm{out} \times d_\mathrm{in}}, \nonumber 
\end{align}
We then apply low-rank factorization to the concatenated matrix  \(W_\mathrm{QKV}\) decomposing it into two low-rank matrices \(B_\mathrm{QKV}\) and \(A_\mathrm{QKV}\) as follows:
\begin{align}
W_\mathrm{QKV} x &\approx B_\mathrm{QKV} A_\mathrm{QKV} x = 
\begin{bmatrix} \hat{W}_\mathrm{Q}^\top & \hat{W}_\mathrm{K}^\top & \hat{W}_\mathrm{V}^\top \end{bmatrix}^\top W_\mathrm{S1} x, \nonumber 
\end{align}
where \(B_\mathrm{QKV} \in \mathbb{R}^{3d_{\mathrm{out}} \times r}\) and \(A_\mathrm{QKV} \in \mathbb{R}^{r \times d_{\mathrm{in}}}\). We then slice \(B_\mathrm{QKV}\) along the output dimension to recover the original shapes corresponding to \(\hat{W}_\mathrm{Q}\), \(\hat{W}_\mathrm{K}\), and \(\hat{W}_\mathrm{V}\) which serve as the reconstruction matrices in the attention module, as shown in Figure~\ref{fig:model}(c). The \(A_\mathrm{QKV}\) corresponds to the shared projection \(W_\mathrm{S1} \in \mathbb{R}^{r \times d_{\mathrm{in}}}\). This approach can also be applied to \textit{gate} and \textit{up} matrices in MLP modules.

With low-rank compression via shared projection, the amortized number of parameters per matrix is reduced to \(r (d_{\mathrm{in}} + C d_{\mathrm{out}}) / C\), and the corresponding FLOPs per matrix become \(2r (d_{\mathrm{in}} + C d_{\mathrm{out}}) / C\), where \(C\) denotes the number of concatenated matrices. We refer to the above method as \textit{Cat}, as illustrated by the orange line in Figure~\ref{fig:flops}. Compared to the naïve approach, \textit{Cat} allows more ranks to be enables the retention of more ranks under the same memory and computational budget.

\begin{figure}[h]
  \centering
  \includegraphics[width=\linewidth]{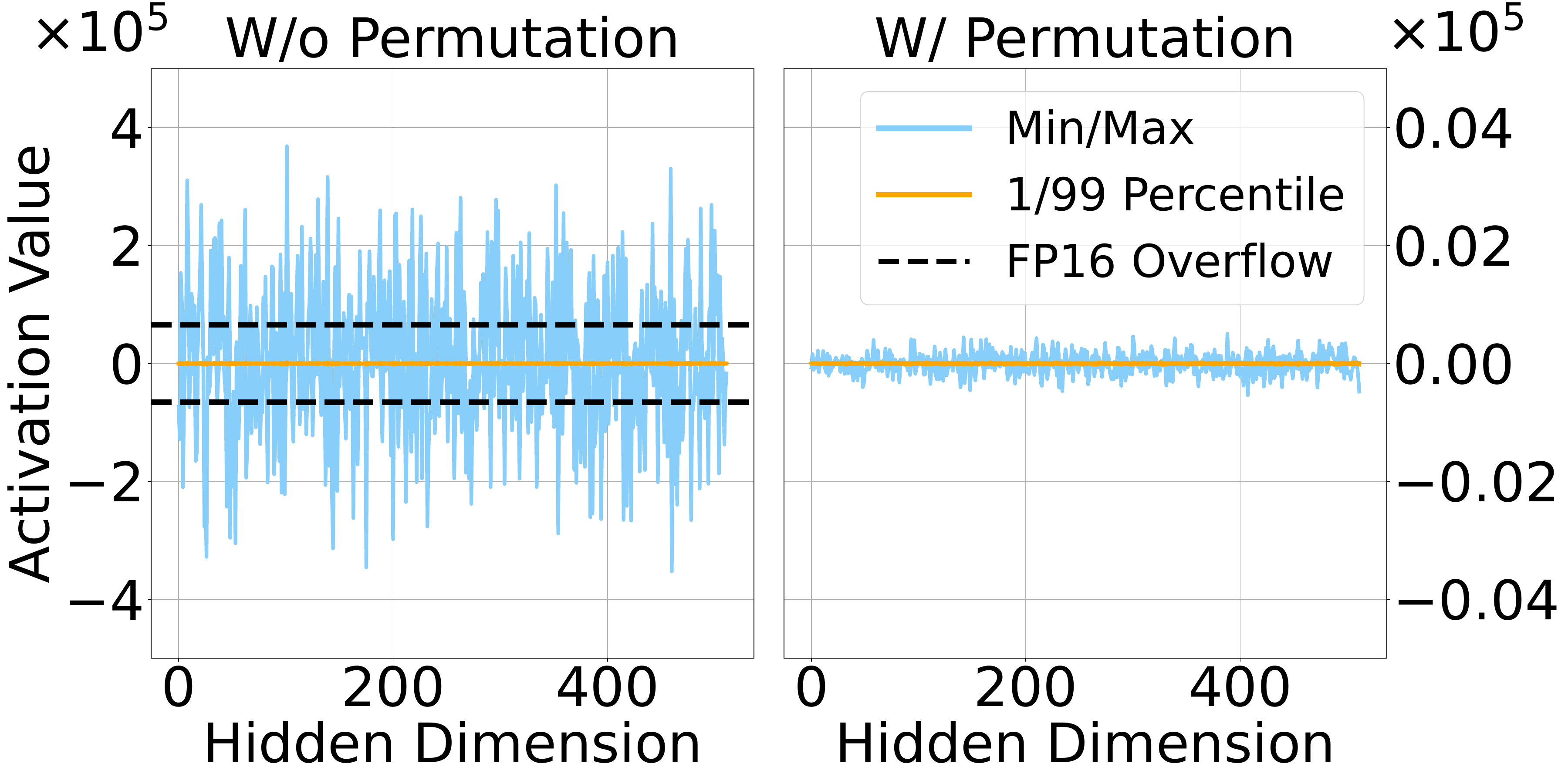}
  \caption{The output activation distribution of low-rank projections in LLaMA2-7B after applying block skipping. Left: Without column permutation, activation outliers cause FP16 overflow. Right: Column permutation stabilizes the distribution, preventing overflow in low-precision inference.}
  \label{fig:act}
\end{figure}

\subsection{Skip: Block Skipping via Schur Complement and Column Permutation}
\label{subsec:skip}

In this subsection, we introduce a novel technique termed Block Skipping (\textit{Skip}), inspired by the Schur complement. Specifically, the multiplication of the weight matrix and the input vector is approximated using two low-rank matrices, a low-rank projection matrix and a reconstruction matrix, as defined in Equation~\eqref{eq:lrapprox}. 
We restructure the low-rank projection matrix into two subblocks and merge the first block into the reconstruction matrix, reducing the overall computational complexity. 
The core idea behind block skipping is to bypass a designated submatrix during computation, significantly reducing cost while maintaining the stability of the computation.

In detail, by analyzing the multiplication of the low-rank projection matrix \( A \in \mathbb{R}^{r \times d_{\mathrm{in}}} \) with the input vector \( x \in \mathbb{R}^{ d_{\mathrm{in}} \times 1 } \). We partition the matrix and vector as:
\[
A = \begin{bmatrix} A_1 & A_2 \end{bmatrix}, \qquad
x^\top = \begin{bmatrix} x_1^\top & x_2^\top \end{bmatrix}^\top
\]
where 
\( A_1 \in \mathbb{R}^{r \times r} \), 
\( A_2 \in \mathbb{R}^{r \times (d_{\mathrm{in}} - r)} \),
\( x_1 \in \mathbb{R}^{r \times 1}\), and
\( x_2 \in \mathbb{R}^{(d_{\mathrm{in}} - r) \times 1}\).
Then, by assuming \( A_1 \) is invertible, the low-rank approximation can be rewritten as:
\begin{align}
W x \approx B A x &= B \begin{bmatrix} A_1 & A_2 \end{bmatrix} \begin{bmatrix} x_1 \\ x_2 \end{bmatrix} \nonumber \\
&= B (A_1 x_1 + A_2 x_2 ) \nonumber \\
&= B A_1 ( x_1 + A_1^{-1} A_2 x_2 ) \nonumber \\
&= B' (x_1 + A' x_2) \label{eq:schur}
\end{align}
by defining \( B' = B A_1 \in \mathbb{R}^{d_{\mathrm{out}} \times r }\) and \( A' = A_1^{-1} A_2 \in \mathbb{R}^{r \times (d_\mathrm{in}-r)}\).

This reformulation eliminates the need to explicitly compute products involving \( A_1 \) by absorbing it into the redefined matrices \( B' \) and \(A'\).
As a result, computing the expression in Equation~\eqref{eq:schur} requires 
\( 2 r (d_{\mathrm{in}} + d_{\mathrm{out}} - r)+ r\) FLOPs, accounting for the multiplication with \(A'\), vector addition, and the final projection via \(B'\). 
The number of parameters is reduced to \( r(d_{\mathrm{in}} + d_{\mathrm{out}} - r) \). 
As shown in Figure~\ref{fig:flops}, \textit{Skip} also retains more ranks than the naïve low-rank compression under the same compression rate.

Despite the promising efficiency of Equation~\eqref{eq:schur}, the formulation may suffer from \textit{numerical instability} when \( A_1 \) is ill-conditioned. In such cases, \( A_1^{-1} \) can have large values, which in turn amplify the magnitude of \( A'\). This amplification can lead to overflow issues when computing \(A' x_2\) under FP16 precision. As shown in the left subfigure of Figure~\ref{fig:act}, although most activation values remain small, some outliers exceed the maximum representable value of FP16, resulting in overflow. Consequently, this prevents inference from being performed reliably under low-precision settings.

To address this, we apply a column permutation \( P \) to the matrix \( A \), such that the leading \( r \) columns of the permuted matrix \( \Tilde{A} = AP = \begin{bmatrix} \Tilde{A}_1 & \Tilde{A}_2 \end{bmatrix} \) form a\textit{ well-conditioned} submatrix \( \Tilde{A}_1 \in \mathbb{R}^{r \times r} \), with the remaining columns denoted by \( \Tilde{A}_2 \in \mathbb{R}^{r \times (d_{\mathrm{in}}-r)} \). 
%
The Strong Rank-Revealing QR factorization~\cite{gu1996rrqr} provides an effective algorithm to identify such well-conditioned column subsets, with provable bounds on condition number. Incorporating this permutation, the original formulation is rewritten as:
\begin{align}
    W x \approx B A x &= B A P P^\top x = B \Tilde{A} \Tilde{x} \nonumber \\
                      &= B \begin{bmatrix} \Tilde{A}_1 \; \Tilde{A}_2 \end{bmatrix} 
                           \begin{bmatrix} \Tilde{x}_1 \\ \Tilde{x}_2 \end{bmatrix} \nonumber \\
                      &= B \Tilde{A}_1 (\Tilde{x}_1 + \Tilde{A}_1^{-1} \Tilde{A}_2 \Tilde{x}_2 ) \nonumber \\
                      &= \Tilde{B}' (\Tilde{x}_1 + \Tilde{A}' \Tilde{x}_2) \label{eq:pschur}
\end{align}
where \( \Tilde{x} = P^{\top} x \), \( \Tilde{B}' = B \Tilde{A}_1 \), and \( \Tilde{A}' = \Tilde{A}_1^{-1} \Tilde{A}_2 \). 

The permutation enhances numerical stability by avoiding inversion of poorly conditioned matrices and improves robustness in low-rank approximation. As shown in the right subfigure of Figure~\ref{fig:act}, our preprocessing step applies a column permutation to the weight matrix, which significantly stabilizes the activation values after computing \( \Tilde{A}' \Tilde{x}_2\). Compared to the left subfigure, the resulting activations are nearly two orders of magnitude smaller and more uniformly distributed. This contribution enables block skipping to be stably executed under FP16 inference without incurring performance degradation.

\begin{table*}[!t]
\resizebox{\textwidth}{!}{%
\begin{tabular}{@{}c|c|cccccccccccc@{}}
\toprule
\multirow{2}{*}{Model} & \multirow{2}{*}{Comp. Rate} & \multicolumn{1}{c|}{\multirow{2}{*}{Method}} & \multicolumn{2}{c|}{Perplexity ↓} & \multicolumn{7}{c}{Zero-shot Task Accuracy (\%) ↑} & Avg. (\%) & Drop (\%) \\
 &  & \multicolumn{1}{c|}{} & Wiki2 & \multicolumn{1}{c|}{C4} & ARC-e & ARC-c & Hella & OBQA & Wino & MathQA & PIQA & ↑ & ↓ \\ \midrule
\multirow{10}{*}{LLaMA2-7B} & - & \multicolumn{1}{c|}{Dense} & 5.47 & \multicolumn{1}{c|}{7.26} & 76.30 & 43.34 & 57.14 & 31.40 & 69.14 & 28.17 & 78.07 & 54.79 & - \\ \cmidrule(l){2-14} 
 & \multirow{5}{*}{20\%} & \multicolumn{1}{c|}{ASVD} & 9.06 & \multicolumn{1}{c|}{11.66} & 67.80 & 34.56 & 48.25 & 29.60 & 63.77 & 25.49 & 72.20 & 48.81 & 5.98 \\
 &  & \multicolumn{1}{c|}{Basis Sharing} & 9.39 & \multicolumn{1}{c|}{23.30} & 59.18 & 27.82 & 38.22 & 24.20 & 66.54 & 23.89 & 66.54 & 43.77 & 11.02 \\
 &  & \multicolumn{1}{c|}{Dobi-SVD} & 9.39 & \multicolumn{1}{c|}{19.46} & 54.25 & 23.29 & 38.80 & 22.80 & 58.56 & 22.75 & 65.34 & 40.83 & 13.97 \\
 &  & \multicolumn{1}{c|}{SVD-LLM} & 8.82 & \multicolumn{1}{c|}{13.42} & 58.67 & 27.65 & 43.10 & 26.20 & 64.25 & 23.85 & 70.18 & 44.84 & 9.95 \\
 &  & \multicolumn{1}{c|}{SkipCat} & \textbf{6.29} & \multicolumn{1}{c|}{\textbf{8.95}} & \textbf{73.23} & \textbf{40.02} & \textbf{51.90} & \textbf{31.00} & \textbf{68.43} & \textbf{27.71} & \textbf{75.84} & \textbf{52.59} & \textbf{2.20} \\ \cmidrule(l){2-14} 
 & \multirow{4}{*}{30\%} & \multicolumn{1}{c|}{ASVD} & 208.55 & \multicolumn{1}{c|}{-} & 34.89 & 23.12 & 29.89 & 13.40 & 51.62 & 22.08 & 56.86 & 33.12 & 21.67 \\
 &  & \multicolumn{1}{c|}{Basis Sharing} & 12.47 & \multicolumn{1}{c|}{38.81} & 51.22 & 22.95 & 34.25 & 19.60 & 59.43 & 23.82 & 61.92 & 39.03 & 15.77 \\
 &  & \multicolumn{1}{c|}{SVD-LLM} & 11.75 & \multicolumn{1}{c|}{19.37} & 52.90 & 24.06 & 38.01 & 22.00 & 61.72 & 22.88 & 67.19 & 41.25 & 13.54 \\
 &  & \multicolumn{1}{c|}{SkipCat} & \textbf{7.65} & \multicolumn{1}{c|}{\textbf{11.57}} & \textbf{68.31} & \textbf{35.67} & \textbf{45.95} & \textbf{26.20} & \textbf{65.98} & \textbf{24.66} & \textbf{72.42} & \textbf{48.46} & \textbf{6.34} \\ \midrule
\multirow{7}{*}{Qwen3-8B} & - & \multicolumn{1}{c|}{Dense} & 9.72 & \multicolumn{1}{c|}{15.42} & 83.54 & 55.97 & 57.13 & 31.00 & 67.80 & 49.61 & 76.88 & 60.28 & - \\ \cmidrule(l){2-14} 
 & \multirow{3}{*}{20\%} & \multicolumn{1}{c|}{ASVD} & 22.10 & \multicolumn{1}{c|}{35.42} & 73.99 & 41.72 & 42.02 & 26.40 & 63.30 & 31.42 & 72.09 & 50.13 & 10.14 \\
 &  & \multicolumn{1}{c|}{SVD-LLM} & 14.33 & \multicolumn{1}{c|}{23.21} & 71.38 & 44.54 & 47.29 & 28.80 & 66.46 & 30.99 & 72.14 & 51.66 & 8.62 \\
 &  & \multicolumn{1}{c|}{SkipCat} & \textbf{11.68} & \multicolumn{1}{c|}{\textbf{19.09}} & \textbf{78.96} & \textbf{49.49} & \textbf{52.47} & \textbf{29.60} & \textbf{69.38} & \textbf{40.37} & \textbf{74.65} & \textbf{56.42} & \textbf{3.86} \\ \cmidrule(l){2-14} 
 & \multirow{3}{*}{30\%} & \multicolumn{1}{c|}{ASVD} & 108.80 & \multicolumn{1}{c|}{138.07} & 47.98 & 23.89 & 30.57 & 17.60 & 50.99 & 23.08 & 61.64 & 36.54 & 23.74 \\
 &  & \multicolumn{1}{c|}{SVD-LLM} & 19.17 & \multicolumn{1}{c|}{32.69} & 58.08 & 33.28 & 40.18 & 23.60 & 61.72 & 25.33 & 68.23 & 44.35 & 15.93 \\
 &  & SkipCat & \textbf{13.81} & \textbf{23.70} & \textbf{74.28} & \textbf{42.41} & \textbf{46.57} & \textbf{28.80} & \textbf{65.43} & \textbf{32.16} & \textbf{72.31} & \textbf{51.71} & \textbf{8.57} \\ \bottomrule
\end{tabular}%
}
\caption{Perplexity and zero-shot accuracy of low-rank compression methods.}
\label{tab:acc}
\end{table*}

\subsection{SkipCat: Efficient Shared Projection with Block Skipping}
\label{subsec:skipcat}
In our work, our primary objective is to increase the number of retained ranks without changing the overall compression ratio. In the preceding subsections, we proposed two methods that individually enhance rank retention under the same budget. Building on these, we further integrate both techniques, as shown in Figure~\ref{fig:model}(d), where all low-rank projections are equipped with block skipping. The benefit of combining both methods can be observed in Figure~\ref{fig:flops}. The purple curve corresponds to applying the \textit{Skip} technique to the low-rank modules. Compared to the standard approach (gray dashed line), our method substantially increases the number of retained ranks in the low-compression regime (i.e., within the $0.75 \sim 1 \times 10^8$ FLOPs or $75 \sim 100$ MB range). While the advantage of preserving more ranks diminishes under higher compression rates, we address this limitation by integrating the \textit{Skip} and \textit{Cat} techniques into a unified approach, referred to as \textit{SkipCat}, as illustrated by the blue line in Figure~\ref{fig:flops}. Our proposed method successfully addresses the limitations of conventional low-rank compression. As illustrated by the blue line in Figure~\ref{fig:flops}, \textit{SkipCat} consistently operates within the effective compression region, demonstrating that the retained ranks contribute proportionally to compression efficiency. This indicates that, under our approach, increasing the number of preserved ranks directly translates into meaningful compression gains. \textit{SkipCat} effectively maximizes the number of effective ranks within a given computational and memory budget.
\section{Experiments}
\subsection{Experimental Setup}
\subsubsection{Models and Datasets}
We evaluate \textit{SkipCat} and previous low-rank compression methods on models from the LLaMA \cite{touvron2023llama} and Qwen \cite{qwen3} families. Since most previous compression methods have been evaluated on LLaMA2, we compare performance using both the 7B and 13B variants. To assess the generalizability of our approach beyond LLaMA-based models, we further evaluate it on Qwen3-8B and Qwen3-14B. We evaluate perplexity on the WikiText-2 \cite{merity2016pointer} and C4 \cite{raffel2020exploring} datasets. In addition, we assess zero-shot task accuracy using the LM-Evaluation-Harness framework \cite{gao2021framework} on a set of benchmark tasks, including ARC-Easy, ARC-Challenge \citep{Clark2018ThinkYH}, HellaSwag \citep{zellers2019hellaswag}, OpenBookQA \citep{OpenBookQA2018}, WinoGrande \citep{sakaguchi2019winogrande}, MathQA \citep{amini2019mathqa} and PIQA \citep{Bisk2020}.

\subsubsection{Experimental Details}
We compare \textit{SkipCat} against several well-known low-rank compression approaches, including ASVD \cite{yuan2023asvd}, Basis Sharing \cite{wang2024basis}, Dobi-SVD \cite{qinsi2025dobisvd}, and SVD-LLM \cite{wang2025svdllm}. Following previous works, our reported compression rates are based on the memory transfer cost associated with the core weight matrices: \textit{q}, \textit{k}, \textit{v}, \textit{o}, \textit{gate}, \textit{up}, and \textit{down}. We use the official codebases of ASVD and Basis Sharing\footnote{Compression rate for Basis Sharing is recalculated using our standard, differing from their original setup.}, and follow their original settings to reproduce the results. We evaluate Dobi-SVD using its officially released “noremapping” model checkpoints without further modification. For SVD-LLM and \textit{SkipCat}, we perform whitening \cite{wang2025svdllm} on the models using samples from the training splits of WikiText-2 and C4.

\begin{table}[H]
\centering
\resizebox{\columnwidth}{!}{%
\begin{tabular}{@{}c|c|c|cc@{}}
\toprule
\multirow{2}{*}{Model}      & \multirow{2}{*}{Comp. Rate} & \multirow{2}{*}{Method} & \multicolumn{2}{c}{Zero-shot Task Accuracy} \\
                            &                             &                         & Avg. (\%) ↑           & Drop (\%) ↓         \\ \midrule
\multirow{5}{*}{LLaMA2-13B} & -                           & Dense                   & 58.07                 & -                   \\ \cmidrule(l){2-5} 
                            & \multirow{2}{*}{20\%}       & SVD-LLM                 & 50.09                 & 7.98                \\
                            &                             & SkipCat                 & \textbf{54.15}        & \textbf{3.92}       \\ \cmidrule(l){2-5} 
                            & \multirow{2}{*}{30\%}       & SVD-LLM                 & 46.15                 & 11.93               \\
                            &                             & SkipCat                 & \textbf{49.29}        & \textbf{8.78}       \\ \midrule
\multirow{5}{*}{Qwen3-14B}  & -                           & Dense                   & 64.01                 & -                   \\ \cmidrule(l){2-5} 
                            & \multirow{2}{*}{20\%}       & SVD-LLM                 & 55.11                 & 8.89                \\
                            &                             & SkipCat                 & \textbf{60.15}        & \textbf{3.85}       \\ \cmidrule(l){2-5} 
                            & \multirow{2}{*}{30\%}       & SVD-LLM                 & 50.19                 & 13.82               \\
                            &                             & SkipCat                 & \textbf{56.04}        & \textbf{7.97}       \\ \bottomrule
\end{tabular}%
}
\caption{Zero-short accuracy of compressed larger models.}
\label{tab:larger_model}
\end{table}

\subsection{Performance Comparison}
To assess overall performance, we compare \textit{SkipCat} with existing low-rank compression methods in terms of perplexity and zero-shot task accuracy, as shown in Table~\ref{tab:acc}. All approaches are evaluated without any additional training or the use of quantization techniques. \textit{SkipCat} outperforms all other methods when evaluated under the same compression rate. \yuchen{On LLaMA2-7B with a 30\% compression rate, our experiments show that existing low-rank compression methods fail to close the performance gap, increasing perplexity on WikiText-2 by 2.1× compared to the uncompressed model, whereas \textit{SkipCat} incurs only a 1.4× increase.} Notably, \textit{SkipCat} at 30\% compression achieves lower perplexity than all other methods compressed at just 20\%. Moreover, in a training-free setting, our method achieves only a 2.2\% average accuracy drop at 20\% compression. When the compression rate increases to 30\%, other methods exhibit a degradation of at least 13.54\%, whereas \textit{SkipCat} limits the drop to 6.34\%, representing a relative improvement of 7\%. The experiments on Qwen3-8B also show substantial improvements in both perplexity and zero-shot task accuracy, further confirming the effectiveness of our approach in compressing models.

To further assess the generalization capability of our method, we extend our experiments to larger models, including LLaMA2-13B and Qwen3-14B. As shown in Table~\ref{tab:larger_model}, despite the increased model scale, \textit{SkipCat} consistently achieves the most effective compression across all methods under the same compression rates.

\newcommand{\cmark}{\ding{51}}  
\newcommand{\xmark}{\ding{55}}  
\begin{table}[t]
\centering
\begin{tabular}{@{}c|ccc|cc@{}}
\toprule
\multirow{2}{*}{Comp. Rate} & \multicolumn{3}{c|}{Ablation Settings}                                & \multicolumn{2}{c}{Perplexity ↓} \\
                            & Cat                   & Skip                  & Quant.                   & Wiki2           & C4             \\ \midrule
\multirow{5}{*}{20\%}       & \xmark & \xmark & \xmark & 8.82            & 13.42          \\
                            & \cmark & \xmark & \xmark & 7.84            & 11.99          \\
                            & \xmark & \cmark & \xmark & 6.71            & 9.32           \\
                            & \cmark & \cmark & \xmark & 6.29            & 8.95           \\
                            & \cmark & \cmark & \cmark & 6.29            & 8.96           \\ \bottomrule
\end{tabular}
\caption{Ablation study results on LLaMA2-7B.}
\label{tab:ablation}
\end{table}

\subsection{Ablation Study}
Our proposed method consists of two core techniques: \textit{Cat} and \textit{Skip}. These two components together form the foundation of \textit{SkipCat}. To examine the individual effectiveness of each technique, we conduct an ablation study on LLaMA2-7B under a 20\% compression rate. As shown in Table~\ref{tab:ablation}, using naïve SVD low-rank compression alone increases the perplexity on WikiText-2 from 5.47 to 8.82. When only the \textit{Cat} technique is applied, the perplexity decreases to 7.84. When only \textit{Skip} is applied, it further reduces to 6.71. These results support our claim in Figure~\ref{fig:flops} that both techniques help increase the number of effective ranks, which in turn reduces performance degradation. When both \textit{Cat} and \textit{Skip} are applied together, the model achieves the lowest perplexity, demonstrating the complementary benefits of combining the two.

We evaluate the compatibility of our approach with quantization. Specifically, we apply HQQ \cite{badri2023hqq} with rescaling to quantize the compressed model to 8-bit precision. The experimental results show that our parameter-level compression can be effectively combined with precision-level quantization. This enables even greater compression without introducing performance loss.

\subsection{Improving Compression with Fine-Tuning}

Our method already shows strong performance compared to existing low-rank compression approaches in a training-free setting. However, in many real-world applications such as domain-specific adaptation or deployment on edge devices, additional fine-tuning can offer further benefits. By combining low-rank compression with task-specific fine-tuning, it is possible not only to improve accuracy on downstream tasks, but also to achieve higher compression rates. To investigate this potential, we fine-tune the low-rank compressed models at various compression levels. Following the setup in SVD-LLM, we apply LoRA-based fine-tuning \cite{hu2022lora} on the Alpaca \cite{taori2023stanford} dataset to adapt the models. 

In this experiment, we use LLaMA2-7B as the base model. The model is first compressed to a target compression ratio, followed by fine-tuning using LoRA. The hyperparameters are set according to the configuration used in SVD-LLM \cite{wang2025svdllm}. As shown in Table~\ref{tab:finetune}, we fine-tune both the \textit{SkipCat}-compressed and SVD-LLM-compressed models and evaluate them on zero-shot tasks. At a 20\% compression rate, we observe that after fine-tuning, the model compressed with \textit{SkipCat} shows only a 0.39\% drop in average accuracy compared to the dense model. In contrast, the model compressed with SVD-LLM still exhibits a 3.78\% accuracy gap. This improvement can be attributed to our method’s ability to maximize the number of effectively utilized ranks through architectural design. By enabling richer capacity within the same compression budget, our approach allows the fine-tuned model to recover performance more effectively than naïve low-rank methods. Even at more aggressive compression rates, our method maintains consistently higher accuracy than SVD-LLM.

\begin{table}[!hb]
\centering
\resizebox{0.8\columnwidth}{!}{%
\begin{tabular}{@{}c|c|cc@{}}
\toprule
\multirow{2}{*}{Comp. Rate} & \multirow{2}{*}{Method} & \multicolumn{2}{c}{Zero-shot Task Accuracy} \\ \cmidrule(l){3-4} 
     &         & Avg. (\%) ↑    & Drop (\%) ↓    \\ \midrule
-    & Dense   & 54.79          & -              \\ \midrule
20\% & SVD-LLM & 51.02          & 3.78           \\
     & SkipCat & \textbf{54.41} & \textbf{0.39}  \\ \midrule
40\% & SVD-LLM & 46.91          & 7.88           \\
     & SkipCat & \textbf{48.65} & \textbf{6.15}  \\ \midrule
60\% & SVD-LLM & 39.50          & 15.29          \\
     & SkipCat & \textbf{41.16} & \textbf{13.64} \\ \midrule
80\% & SVD-LLM & 32.33          & 22.47          \\
     & SkipCat & \textbf{32.82} & \textbf{21.97} \\ \bottomrule
\end{tabular}%
}
\caption{Zero-shot accuracy of fine-tuned compressed models based on LLaMA2-7B.}
\label{tab:finetune}
\end{table}



\section{Conclusion}
In this work, we propose \textit{SkipCat}, a novel low-rank compression framework for large language models. Our method integrates two key techniques: \textit{Cat} and \textit{Skip}. The former allows multiple matrices within a single layer to share a common low-rank projection, while the latter, block skipping, reduces computation by omitting calculations in certain sub-blocks of the matrices. These components allow the model to increase the number of effectively utilized ranks without changing the overall compression ratio, which helps maximize performance. We conduct evaluations on downstream tasks and compare \textit{SkipCat} with existing low-rank compression methods. Our method consistently achieves significantly higher accuracy under the same compression rates. These results demonstrate that increasing the number of effective ranks is a highly effective strategy for improving the quality of compressed models.

\bibliography{aaai2026}

\clearpage

\appendix
\setcounter{secnumdepth}{2} 
\onecolumn
\section*{Supplementary Material}

\section{The Relation between Block Skipping and Schur Complement}

Considering a full rank weight matrix with  SVD decomposition $W = B_{\mathrm{full}}A_{\mathrm{full}} \in \mathbb{R}^{d_\mathrm{out} \times d_\mathrm{in}}$.
The na\"ive low-rank compression splits each of the matrix $B_{\mathrm{full}} \in \mathbb{R}^{d_\mathrm{out} \times d_\mathrm{in}}$ and $A_{\mathrm{full}}\in \mathbb{R}^{d_\mathrm{in} \times d_\mathrm{in}}$ into 2 subblocks. One subblock contains the necessary leading basis, and the other block contains the unnecessary basis. For example, $B_{\mathrm{full}} = \begin{bmatrix} B_{1} & B_{2} \end{bmatrix}$. The first subblock $B_{1} \in \mathbb{R}^{d_\mathrm{out} \times r}$ contains the leading \(r\) left singular vectors scaled by the square root of their singular values, and $B_{2} \in \mathbb{R}^{d_\mathrm{out} \times (d_\mathrm{in}-r)}$ are the remaining $d_\mathrm{in}-r$ components.

On the other hand, the proposed Block Skipping considering not only the leading ranks, but also more detailed structure of each singular vector.
The matrix $A_{\mathrm{full}}\in \mathbb{R}^{d_\mathrm{in} \times d_\mathrm{in}}$ is split into 4 subblocks. Two of them are disjoint dimension of the leading $r$ basis, and others are disjoint dimension of the remaining basis. That is,
\begin{equation}
    A_{\mathrm{full}} = \begin{bmatrix} A_{11} & A_{12} \\ A_{21} & A_{22} \end{bmatrix}
\end{equation}
where $A_{11} \in \mathbb{R}^{r \times r}$ and $A_{12} \in \mathbb{R}^{r \times (d_\mathrm{in}-r)}$ are two disjoint dimension of the leading $r$ right singular vectors scaled by the square root of their singular values, while $A_{21} \in \mathbb{R}^{(d_\mathrm{in}-r) \times r}$ and $A_{22} \in \mathbb{R}^{(d_\mathrm{in}-r) \times (d_\mathrm{in}-r)}$ are the remaining $d_\mathrm{in}-r$ vectors.

By applying the Schur complement, the matrix $A_{\mathrm{full}}$ is decomposed into
\begin{align}
    A_{\mathrm{full}} 
    &= \begin{bmatrix} A_{11} & A_{12} \\ A_{21} & A_{22} \end{bmatrix} \nonumber \\
    &= \begin{bmatrix}
        I_r & 0 \\ A_{21}A_{11}^{-1} & I_{d_\mathrm{in}}
    \end{bmatrix}
    \begin{bmatrix}
        A_{11} & 0 \\ 0 & D
    \end{bmatrix}
    \begin{bmatrix}
        I_{r} & A_{11}^{-1} A_{12} \\ 0 & I_{d_\mathrm{in}}
    \end{bmatrix}
\end{align}
where $D=A_{22}-A_{21}A_{11}^{-1}A_{12}$.

Merge the first two matrices of the decomposition into the matrix $B_{\mathrm{full}}$.
\begin{align}
    B_\mathrm{new} &= B_{\mathrm{full}} \begin{bmatrix}
        I_r & 0 \\ A_{21}A_{11}^{-1} & I_{d_\mathrm{in}}
    \end{bmatrix}
    \begin{bmatrix}
        A_{11} & 0 \\ 0 & D
    \end{bmatrix} \nonumber \\
    &= \begin{bmatrix}
        B_{1}A_{11}+B_{2}A_{21} & B_{2}D \\
    \end{bmatrix} \\
    A_\mathrm{new} &= \begin{bmatrix}
        I_{r} & A_{11}^{-1} A_{12} \\ 0 & I_{d_\mathrm{in}}
    \end{bmatrix}
\end{align}


Consequently, with considering the low-rank approximation with rank-$r$, which means the matrices $B_{2}$, $A_{21}$, and $A_{22}$ are ignored. It becomes
\begin{align}
    Wx &= B_\mathrm{new}A_\mathrm{new}x \nonumber\\
       &= \begin{bmatrix}
           B_{1}A_{11} & 0
       \end{bmatrix}
       \begin{bmatrix}
           I_r & A_{11}^{-1}A_{12} \\ 0 & I_{d_\mathrm{in}}
       \end{bmatrix} \begin{bmatrix}
           x_1 \\ x_2
       \end{bmatrix} \nonumber\\
       &= \begin{bmatrix}
           B_{1}A_{11} & 0
       \end{bmatrix} 
       \begin{bmatrix}
       x_1 + A_{11}^{-1}A_{12}x_2 \\ x_2
       \end{bmatrix} \nonumber\\
       &= B_{1}A_{11} (x_1 + A_{11}^{-1}A_{12}x_2) \nonumber\\
       &= B' (x_1 + A' x_2)
\end{align}
where $B'=B_{1}A_{11}$ and $A'=A_{11}^{-1}A_{12}$, and it is the same result as in Equation~\eqref{eq:schur}.
\section{Shared Low-rank Projection for MLP Module}
In this section, we provide additional details on how the shared projection is constructed within the MLP module. In this case, the two projection matrices, \(W_\mathrm{G}\) and \(W_\mathrm{U}\), receive the same input. To exploit this structure, we concatenate the matrices along the output dimension:
\begin{align}
W_\mathrm{GU} = \begin{bmatrix} W_\mathrm{G}^\top & W_\mathrm{U}^\top \end{bmatrix}^\top \in \mathbb{R}^{2d_\mathrm{out} \times d_\mathrm{in}}. \nonumber
\end{align}

We then perform low-rank decomposition on the concatenated matrix to obtain:
\begin{align}
W_\mathrm{GU} \approx B_\mathrm{GU} A_\mathrm{GU} = 
\begin{bmatrix} \hat{W}_\mathrm{G}^\top & \hat{W}_\mathrm{U}^\top \end{bmatrix}^\top W_\mathrm{S2}, \nonumber
\end{align}
where \(B_\mathrm{GU} \in \mathbb{R}^{2d_\mathrm{out} \times r}\) is the reconstruction matrix, and \(A_\mathrm{GU} = W_\mathrm{S2} \in \mathbb{R}^{r \times d_\mathrm{in}}\) is the shared projection matrix.

By splitting \(B_\mathrm{GU}\) according to the original output dimensions, we recover the individual reconstruction matrices \(\hat{W}_\mathrm{G}\) and \(\hat{W}_\mathrm{U}\), as used in the MLP module.

The resulting approximation takes the form:
\begin{align}
W_\mathrm{GU} x \approx B_\mathrm{GU} W_\mathrm{S2} x. \nonumber
\end{align}

Under this construction, the number of parameters becomes \(r(d_{\mathrm{in}} + C d_{\mathrm{out}})\), and the total FLOPs is \(2r(d_{\mathrm{in}} + C d_{\mathrm{out}})\), where \(C = 2\) in this case. This design applies the same principle used in the attention module and provides a consistent means to increase rank capacity without increasing computational cost.

\section{Numerical Stability of Block Skipping}
\begin{figure}[h]
  \centering
  \begin{subfigure}{0.48\linewidth}
    \centering
    \includegraphics[width=\linewidth]{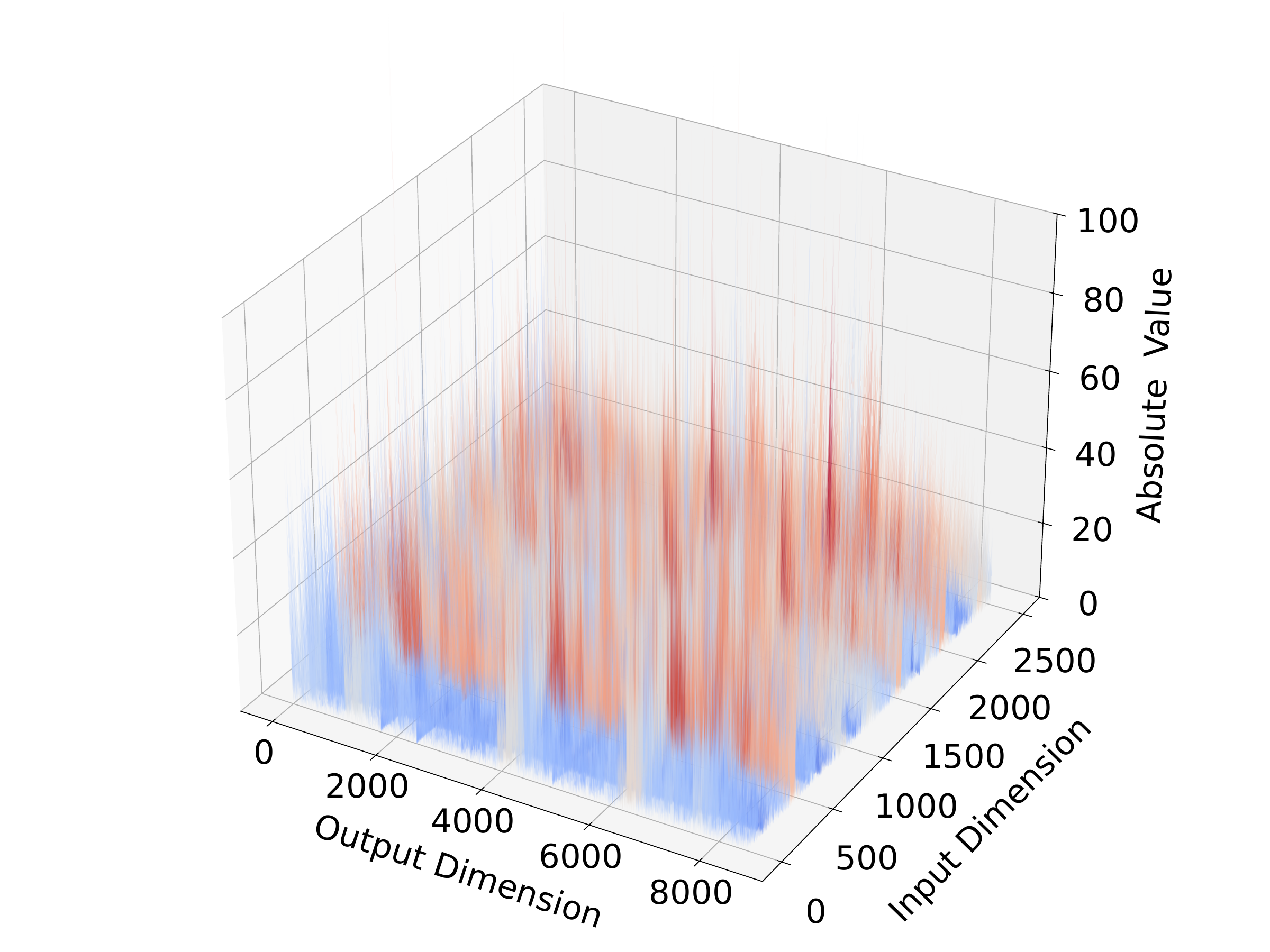}
    \caption{W/o column permutation}
    \label{fig:weight_noP}
  \end{subfigure}
  \hfill
  \begin{subfigure}{0.48\linewidth}
    \centering
    \includegraphics[width=\linewidth]{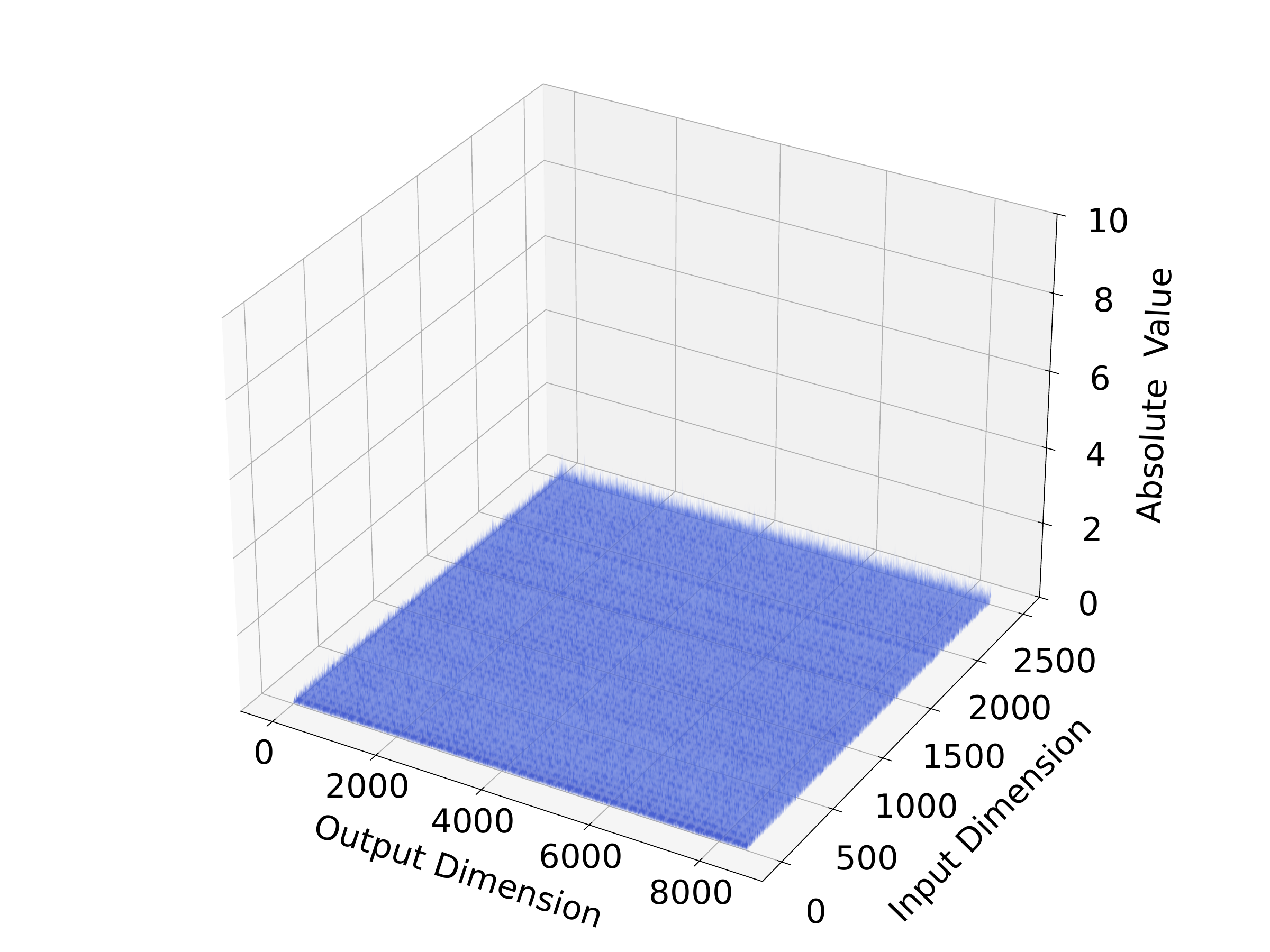}
    \caption{W/ column permutation}
    \label{fig:weight_P}
  \end{subfigure}
  \caption{Visualization comparing the weights after applying Block Skipping, with and without column permutation. Applying column permutation before Block Skipping reduces outliers and improves numerical stability.}
  \label{fig:weight}
\end{figure}

In Section~\ref{subsec:skip}, we introduced the Block Skipping technique and noted that directly applying it may lead to numerical instability, often resulting in overflow during FP16 computation. One of our key contributions is the introduction of column permutation to improve numerical stability. Figure~\ref{fig:weight_noP} visualizes a low-rank projection matrix in LLaMA2-7B after applying Block Skipping without column permutation. As shown, this leads to a large number of unusually high outliers. When such weights are multiplied by activation values that also contain outliers, FP16 arithmetic is prone to overflow.
In contrast, Figure~\ref{fig:weight_P} shows the result of applying column permutation before Block Skipping. The resulting weight values are significantly more stable and smaller in magnitude, which helps prevent overflow during FP16 model inference.


We also conduct experiments using BF16 for compressed model inference to examine whether it can mitigate the overflow issue. As shown in Table~\ref{tab:datatype}, we evaluate the perplexity on the WikiText-2 dataset using LLaMA2-7B compressed with Block Skipping, without applying column permutation, across different data types.
Since FP32 has a sufficiently large numerical range, the model can still operate correctly even after applying Block Skipping. Although BF16 does not cause overflow, its limited precision results in a significant degradation in perplexity. This suggests that simply adopting BF16 is insufficient to resolve the numerical instability issue. In contrast, our approach, which applies column permutation to the weights before Block Skipping, provides a more reliable and stable solution.

\begin{table}[h]
\centering
\resizebox{0.35\columnwidth}{!}{%
\begin{tabular}{@{}c|c|c@{}}
\toprule
Comp. Rate            & Data Type & Wiki2 Perplexity ↓ \\ \midrule
\multirow{3}{*}{10\%} & FP32      & 5.68        \\
                      & FP16      & 13.34       \\
                      & BF16      & 31628.76    \\ \midrule
\multirow{3}{*}{20\%} & FP32      & 6.19        \\
                      & FP16      & Overflow         \\
                      & BF16      & 11294.57    \\ \midrule
\multirow{3}{*}{30\%} & FP32      & 7.27        \\
                      & FP16      & Overflow         \\
                      & BF16      & 82674.39    \\ \bottomrule
\end{tabular}%
}
\caption{Comparison of perplexity across different data types for LLaMA2-7B with Block Skipping applied without prior column permutation.}
\label{tab:datatype}
\end{table}
\section{Stabilizing Quantization for Block Skipping}
While we applied column permutation to allow model stable inference under FP16, there is still a slight degradation in perplexity when further reducing the precision to 8-bit, as shown in Table~\ref{tab:ablation}. Although column permutation significantly reduces outliers in the weights, this method alone is insufficient for stable low-precision quantization. We mitigate this quantization loss by applying a Hadamard matrix \cite{tseng2024quip, ashkboos2024quarot} and channel scaling \cite{neyshabur2018towards, meller2019same} across low-rank projection and reconstruction matrices to reduce and equalize the quantization difficulty between the low-rank matrices before quantization, enabling stable 8-bit quantization results.

As shown in Table~\ref{tab:quant_ablation}, we perform an ablation study by comparing the perplexity of LLaMA2-7B compressed by \textit{SkipCat} at a 20\% compression rate to demonstrate the effectiveness of the Hadamard transform and channel scaling. When both methods are applied, the 8-bit quantized model achieves perplexity on WikiText-2 comparable to its full-precision counterpart.

\begin{table}[h]
\centering
\resizebox{0.45\linewidth}{!}{%
\begin{tabular}{@{}c|ccc|cc@{}}
\toprule
\multirow{2}{*}{Comp. Rate} & \multicolumn{3}{c|}{Ablation Settings}                                & \multicolumn{2}{c}{Perplexity ↓} \\
                            & Quant                 & Hadamard              & Scaling               & Wiki2           & C4             \\ \midrule
\multirow{5}{*}{20\%}       & \xmark & \xmark & \xmark & 6.29            & 8.95           \\
                            & \cmark & \xmark & \xmark & 6.36            & 9.06           \\
                            & \cmark & \cmark & \xmark & 6.32            & 8.97           \\
                            & \cmark & \xmark & \cmark & 6.30             & 8.98           \\
                            & \cmark & \cmark & \cmark & 6.29            & 8.96           \\ \bottomrule
\end{tabular}%
}
\caption{Ablation study on quantization stabilization.}
\label{tab:quant_ablation}
\end{table}
\section{Calibration Strategy for Whitening}
\begin{figure*}[h]
  \centering
  \includegraphics[width=0.9\textwidth]{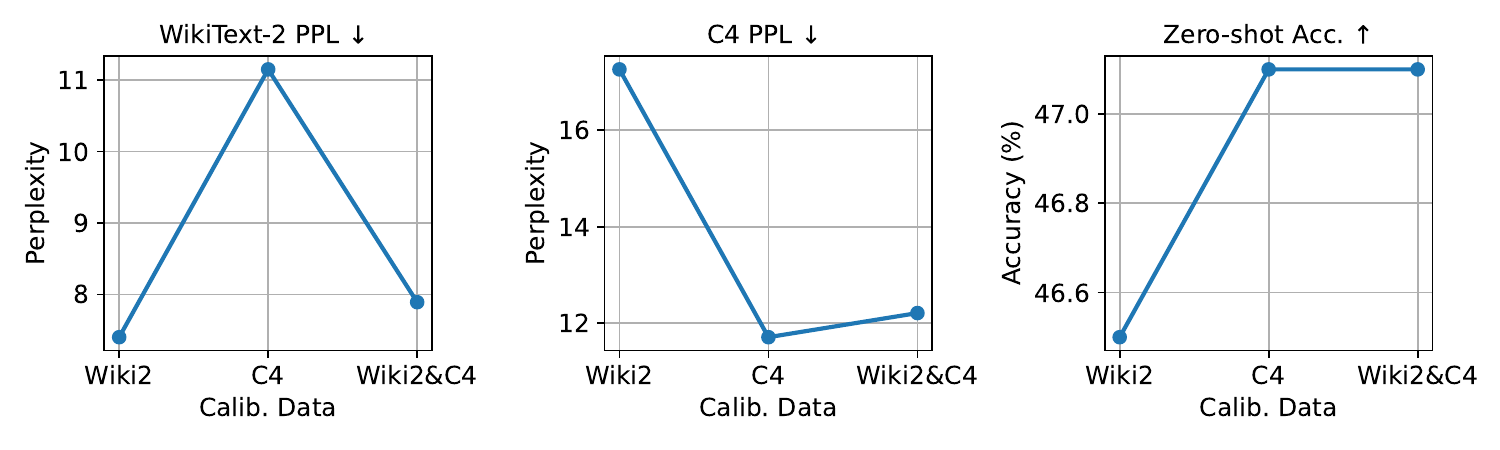}
  \caption{Perplexity and zero-shot accuracy of compressed LLaMA2-7B using different calibration datasets for whitening.}
  \label{fig:calib_data}
\end{figure*}

Before applying low-rank compression to the model, we first perform weight whitening \cite{wang2025svdllm}. This whitening step requires a calibration dataset. To determine an appropriate dataset, we conduct an experiment comparing several options. Specifically, we consider three types of calibration data: (1) 256 sequences of 2048 tokens sampled from the WikiText-2 training set, (2) 256 sequences from the C4 training set, and (3) a mixed dataset composed of 128 sequences from each of WikiText-2 and C4. Using LLaMA2-7B, we apply whitening with each of the three calibration datasets, followed by \textit{SkipCat} compression at a 20\% compression rate. We then evaluate the resulting models in terms of perplexity on WikiText-2 and C4, as well as the average accuracy on zero-shot tasks. As shown in Figure~\ref{fig:calib_data}, the mixed dataset of WikiText-2 and C4 yields the most consistent performance across all three benchmarks. Therefore, we adopt this mixed dataset as the calibration set for whitening in our \textit{SkipCat} framework.

\begin{figure}[h]
  \centering
  \includegraphics[width=0.5\columnwidth]{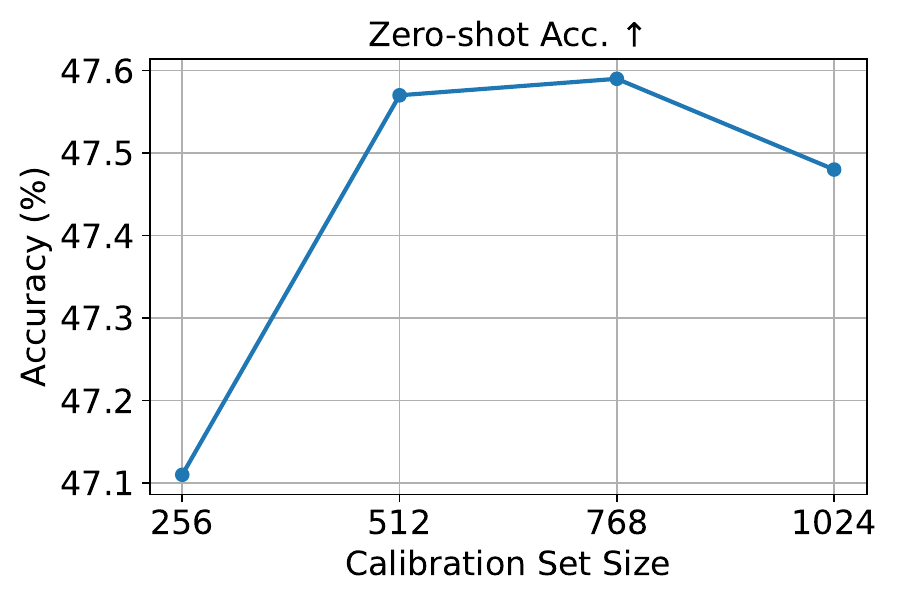}
  \caption{Effect of calibration dataset size on zero-shot accuracy after whitening.}
  \label{fig:calib_amount}
\end{figure}

We also conduct an experiment to determine the appropriate number of calibration samples. As shown in Figure~\ref{fig:calib_amount}, we vary the number of mixed calibration samples and apply whitening to LLaMA2-7B, followed by 20\% compression using \textit{SkipCat}. We then evaluate the average accuracy on zero-shot tasks.The results show that using 512 calibration samples is sufficient to achieve stable performance. Therefore, in our \textit{SkipCat} framework, we adopt 512 mixed calibration samples for model whitening.

\section{Supplementary Results}

In this section, we provide the full zero-shot evaluation results on LLaMA2-13B and Qwen3-14B, as shown in Table~\ref{tab:larger_full}. It can be observed that \textit{SkipCat} consistently outperforms previous methods across all tasks. Additionally, we include the complete results of LoRA fine-tuning applied to the compressed LLaMA2-7B, with evaluations conducted across various tasks. These results are summarized in Table~\ref{tab:finetune_full}.

\begin{table*}[h]
\centering
\resizebox{\textwidth}{!}{%
\begin{tabular}{@{}c|c|c|cc|ccccccccc@{}}
\toprule
\multirow{2}{*}{Model} & \multirow{2}{*}{Comp. Rate} & \multirow{2}{*}{Method} & \multicolumn{2}{c|}{Perplexity ↓} & \multicolumn{7}{c}{Zero-shot Task Accuracy (\%) ↑} & Avg. (\%) & Drop (\%) \\
 &  &  & Wiki2 & C4 & ARC-e & ARC-c & Hella & OBQA & Wino & MathQA & PIQA & ↑ & ↓ \\ \midrule
\multirow{5}{*}{LLaMA2-13B} & - & Dense & 4.88 & 6.73 & 79.42 & 48.29 & 60.07 & 35.20 & 72.22 & 32.19 & 79.11 & 58.07 & - \\ \cmidrule(l){2-14} 
 & \multirow{2}{*}{20\%} & SVD-LLM & 6.83 & 10.80 & 69.49 & 36.18 & 47.63 & 28.80 & 68.11 & 25.29 & 75.14 & 50.09 & 7.98 \\
 &  & SkipCat & \textbf{5.79} & \textbf{8.50} & \textbf{74.96} & \textbf{39.93} & \textbf{53.72} & \textbf{32.20} & \textbf{72.14} & \textbf{28.68} & \textbf{77.42} & \textbf{54.15} & \textbf{3.92} \\ \cmidrule(l){2-14} 
 & \multirow{2}{*}{30\%} & SVD-LLM & 8.43 & 14.39 & 63.51 & 29.18 & 42.03 & 25.60 & 67.32 & 24.05 & 71.33 & 46.15 & 11.93 \\
 &  & SkipCat & \textbf{6.99} & \textbf{11.19} & \textbf{67.42} & \textbf{33.79} & \textbf{45.78} & \textbf{27.80} & \textbf{69.69} & \textbf{25.76} & \textbf{74.81} & \textbf{49.29} & \textbf{8.78} \\ \midrule
\multirow{5}{*}{Qwen3-14B} & - & Dense & 8.64 & 13.81 & 84.13 & 58.62 & 60.96 & 34.80 & 73.01 & 56.45 & 80.09 & 64.01 & - \\ \cmidrule(l){2-14} 
 & \multirow{2}{*}{20\%} & SVD-LLM & 12.38 & 20.72 & 75.08 & 47.35 & 51.03 & 31.40 & 70.80 & 35.11 & 75.03 & 55.11 & 8.89 \\
 &  & SkipCat & \textbf{10.33} & \textbf{16.62} & \textbf{81.36} & \textbf{53.75} & \textbf{56.67} & \textbf{33.80} & \textbf{71.43} & \textbf{46.87} & \textbf{77.20} & \textbf{60.15} & \textbf{3.85} \\ \cmidrule(l){2-14} 
 & \multirow{2}{*}{30\%} & SVD-LLM & 15.13 & 27.39 & 70.20 & 39.16 & 44.80 & 28.60 & 66.93 & 30.05 & 71.60 & 50.19 & 13.82 \\
 &  & SkipCat & \textbf{12.73} & \textbf{21.26} & \textbf{76.60} & \textbf{47.35} & \textbf{51.76} & \textbf{33.00} & \textbf{70.56} & \textbf{38.12} & \textbf{74.86} & \textbf{56.04} & \textbf{7.97} \\ \bottomrule
\end{tabular}%
}
\caption{Perplexity and zero-shot accuracy of compressed larger models.}
\label{tab:larger_full}
\end{table*}

\begin{table*}[!htbp]
\centering
\resizebox{0.8\textwidth}{!}{%
\begin{tabular}{@{}c|c|ccccccccc@{}}
\toprule
\multirow{2}{*}{Comp. Rate} & \multirow{2}{*}{Method} & \multicolumn{7}{c}{Zero-shot Task Accuracy (\%) ↑}                                                                   & Avg. (\%)      & Drop (\%)      \\
                            &                         & ARC-e          & ARC-c          & Hella          & OBQA           & Wino           & MathQA         & PIQA           & ↑              & ↓              \\ \midrule
-                           & Dense                   & 76.30          & 43.34          & 57.14          & 31.40          & 69.14          & 28.17          & 78.07          & 54.79          & -              \\ \midrule
20\%                        & SVD-LLM                 & 70.03          & 38.23          & 52.88          & 29.60          & 65.19          & 25.29          & 75.90          & 51.02          & 3.78           \\
                            & SkipCat                 & \textbf{75.08} & \textbf{42.83} & \textbf{56.00} & \textbf{33.80} & \textbf{68.11} & \textbf{27.54} & \textbf{77.48} & \textbf{54.41} & \textbf{0.39}  \\ \midrule
40\%                        & SVD-LLM                 & 63.76          & 33.11          & 47.05          & 26.40          & 61.96          & 23.95          & 72.14          & 46.91          & 7.88           \\
                            & SkipCat                 & \textbf{66.71} & \textbf{33.62} & \textbf{49.92} & \textbf{28.60} & \textbf{62.27} & \textbf{24.66} & \textbf{74.76} & \textbf{48.65} & \textbf{6.15}  \\ \midrule
60\%                        & SVD-LLM                 & 50.29          & 22.87          & 37.96          & 21.80          & 55.72          & \textbf{23.15} & 64.74          & 39.50          & 15.29          \\
                            & SkipCat                 & \textbf{53.28} & \textbf{26.28} & \textbf{40.00} & \textbf{23.20} & \textbf{57.38} & 22.41          & \textbf{65.56} & \textbf{41.16} & \textbf{13.64} \\ \midrule
80\%                        & SVD-LLM                 & 35.86          & 19.88          & 29.28          & 11.40          & 50.04          & \textbf{22.11} & \textbf{57.73} & 32.33          & 22.47          \\
                            & SkipCat                 & \textbf{36.11} & \textbf{20.31} & \textbf{29.29} & \textbf{14.6}  & \textbf{51.3}  & 21.14          & 57.02          & \textbf{32.82} & \textbf{21.97} \\ \bottomrule
\end{tabular}%
}
\caption{Zero-shot accuracy of fine-tuned compressed models.}
\label{tab:finetune_full}
\end{table*}

In addition to the aforementioned models, we also evaluated our low-rank compression method on LLaMA-3.1-8B using zero-shot tasks, as reported in Table~\ref{tab:llama31}. The results demonstrate that our method consistently achieves competitive performance.

\begin{table}[]
\centering
\resizebox{0.8\textwidth}{!}{%
\begin{tabular}{@{}ccccccccccc@{}}
\toprule
\multirow{2}{*}{Comp. Rate} & \multirow{2}{*}{Method} & \multicolumn{7}{c}{Zero-shot Task Accuracy (\%) $\uparrow$}                                                          & Avg. (\%)      & Drop (\%)      \\
                            &                         & ARC-e          & ARC-c          & Hella          & OBQA           & Wino           & MathQA         & PIQA           & $\uparrow$     & $\downarrow$   \\ \midrule
-                           & Dense                   & 81.48          & 51.19          & 60.01          & 33.40          & 73.64          & 39.60          & 80.25          & 59.94          & -              \\ \midrule
20\%                        & SVD-LLM                 & 58.80          & 28.58          & 41.40          & 19.60          & 61.40          & 25.06          & 68.99          & 43.40          & 16.53          \\
                            & SkipCat                 & \textbf{68.90} & \textbf{36.60} & \textbf{50.13} & \textbf{25.60} & \textbf{65.67} & \textbf{29.82} & \textbf{74.86} & \textbf{50.23} & \textbf{9.71}  \\ \midrule
30\%                        & SVD-LLM                 & 45.58          & 22.18          & 33.36          & 17.60          & 56.20          & 23.55          & 62.19          & 37.24          & 22.70          \\
                            & SkipCat                 & \textbf{51.22} & \textbf{23.55} & \textbf{38.60} & \textbf{19.00} & \textbf{56.91} & \textbf{24.49} & \textbf{65.45} & \textbf{39.89} & \textbf{20.05} \\ \bottomrule
\end{tabular}%
}
\caption{Zero-shot accuracy of compressed LLaMA3.1-8B.}
\label{tab:llama31}
\end{table}

Among existing low-rank compression methods, SVD-LLM V2 \cite{wang2025svd} is a well-known approach. However, as its official source code has not yet been released, conducting a direct comparison under identical experimental conditions remains challenging. In the original SVD-LLM V2 paper, the authors reported a zero-shot accuracy of 54\% on LLaMA-7B at a 20\% compression rate. To provide a meaningful comparison, we also evaluated our method on the same model under the same compression setting. As shown in Table~\ref{tab:svdllmv2}, our method achieves a higher zero-shot accuracy of 56\% at the same compression rate. Note that the results for SVD-LLM V2 presented in Table~\ref{tab:svdllmv2} are directly taken from its original paper.

\begin{table}[h]
\centering
\resizebox{0.8\textwidth}{!}{%
\begin{tabular}{@{}cccccccccc@{}}
\toprule
\multirow{2}{*}{Comp. Rate} & \multirow{2}{*}{Method} & \multicolumn{6}{c}{Zero-shot Task Accuracy (\%) $\uparrow$}                       & Avg. (\%)   & Drop (\%)    \\
                            &                         & ARC-e       & Hella       & OBQA        & Wino        & MathQA      & PIQA        & $\uparrow$  & $\downarrow$ \\ \midrule
-                           & Dense                   & 75          & 57          & 34          & 70          & 27          & 79          & 57          & -            \\ \midrule
\multirow{2}{*}{20\%}       & SVD-LLM V2              & 72          & 52          & 32          & \textbf{70} & 24          & 75          & 54          & 3            \\
                            & SkipCat                 & \textbf{75} & \textbf{56} & \textbf{35} & 68          & \textbf{26} & \textbf{78} & \textbf{56} & \textbf{1}   \\ \bottomrule
\end{tabular}%
}
\caption{Comparison of zero-shot accuracy between SVD-LLM V2 and our method on LLaMA-7B at a 20\% compression rate. The results for SVD-LLM V2 are taken from the original paper.}
\label{tab:svdllmv2}
\end{table}
\section{Inference Efficiency and Throughput Analysis}
\begin{figure}[h]
  \centering
  \includegraphics[width=0.5\columnwidth]{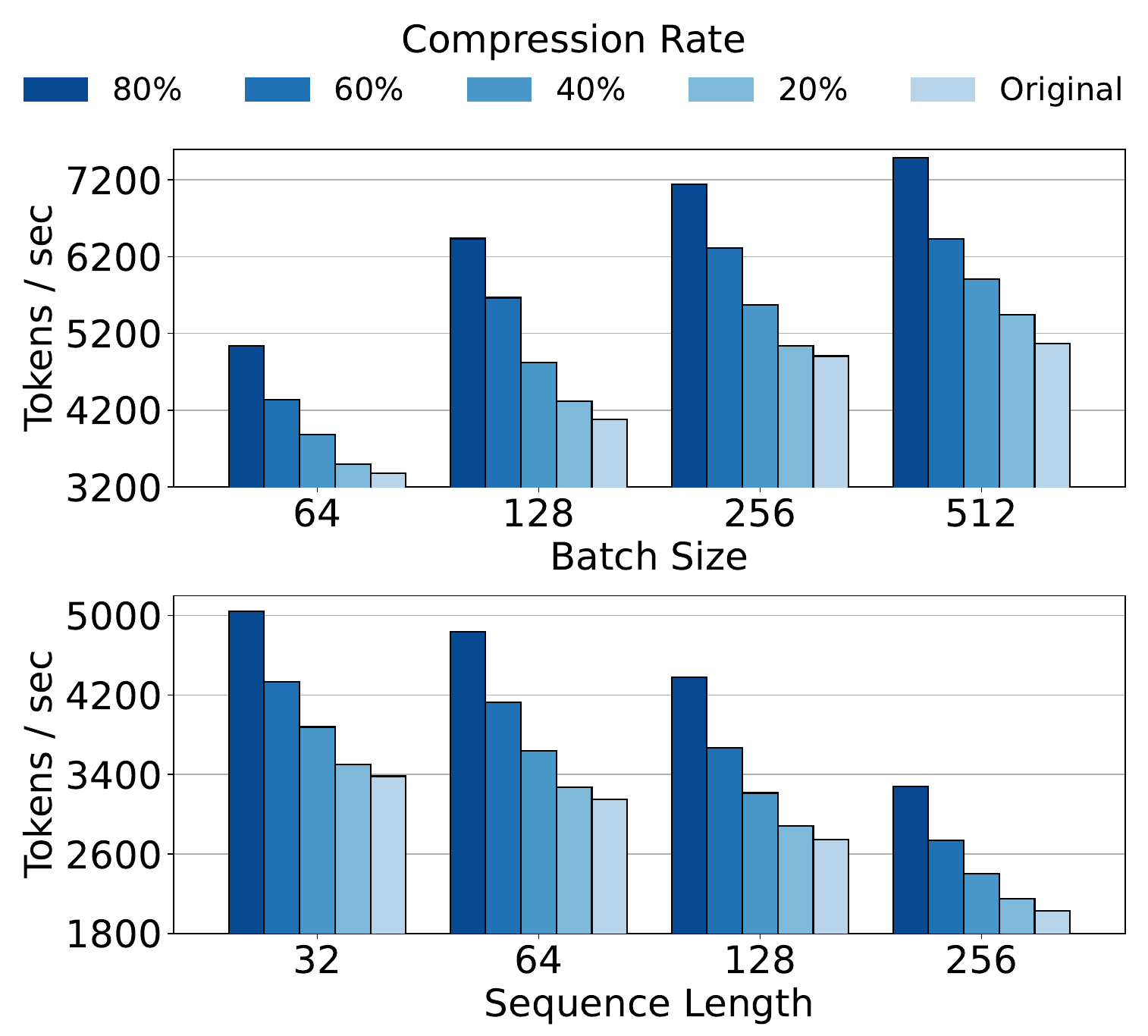}
  \caption{Throughput of LLaMA2-7B under different compression rates with varying batch sizes and input sequence lengths. Top: Throughput results with a fixed input sequence length of 32 and varying batch sizes. Bottom: Throughput results with a fixed batch size of 64 and varying input sequence lengths.}
  \label{fig:speed}
\end{figure}

\textit{SkipCat} reduces the FLOPs of the model, making it suitable for evaluating the actual inference latency during the prefilling stage of LLMs. As shown in Table~\ref{tab:ttft}, we measure the Time to First Token (TTFT) on an A100 GPU under various compression rates. The results demonstrate that \textit{SkipCat} not only reduces memory usage but also improves the prefilling latency. Figure~\ref{fig:speed} presents the throughput evaluation under different compression rates. Across various settings, \textit{SkipCat} consistently improves the decoding speed of the model, demonstrating its effectiveness in accelerating the decoding stage.
\begin{table}[h]
\centering
\resizebox{0.4\columnwidth}{!}{%
\begin{tabular}{@{}c|c|c|c@{}}
\toprule
Method & Comp. Rate & TTFT (sec) & Speedup \\ \midrule
Dense & - & 1.14 & - \\ \midrule
\multirow{4}{*}{SkipCat} & 20\% & 0.99 & 1.15x \\
 & 40\% & 0.79 & 1.44x \\
 & 60\% & 0.57 & 2.00x \\
 & 80\% & 0.35 & 3.26x \\ \bottomrule
\end{tabular}%
}
\caption{Time to First Token (TTFT) of LLaMA2-7B with SkipCat compression at varying compression rates.}
\label{tab:ttft}
\end{table}

We believe that \textit{SkipCat} can also reduce the memory transfer time for model parameters, making it well-suited for model offloading scenarios. When deploying a model on a GPU with limited VRAM (e.g., less than 8GB), a portion of the model's parameters must be offloaded to host DRAM and transferred back to GPU VRAM only when needed. In such cases, memory transfer becomes a major bottleneck.
\textit{SkipCat} can help alleviate this issue by reducing the size of the parameters that need to be transferred. To evaluate this, we offload half of the model layers to DRAM and measure the throughput after applying \textit{SkipCat} compression. As shown in Table~\ref{tab:speed_offload}, \textit{SkipCat} significantly improves performance in the offloading setting. These results suggest that low-rank compression is a promising approach for efficient model offloading.
\begin{table}[!t]
\centering
\resizebox{0.4\columnwidth}{!}{%
\begin{tabular}{@{}c|c|c|c@{}}
\toprule
Method & Comp. Rate & \begin{tabular}[c]{@{}c@{}}Throughput \\ (tokens/sec)\end{tabular} & Speedup \\ \midrule
Dense & - & 8.63 & - \\ \midrule
\multirow{4}{*}{SkipCat} & 20\% & 16.10 & 1.87x \\
 & 40\% & 20.84 & 2.41x \\
 & 60\% & 33.99 & 3.94x \\
 & 80\% & 58.05 & 6.73x \\ \bottomrule
\end{tabular}%
}
\caption{Throughput of LLaMA2-7B under model offloading with varying SkipCat compression rates.}
\label{tab:speed_offload}
\end{table}
\end{document}